\documentclass[10pt, a4paper]{article}

\usepackage{lrec-coling2024} 
\usepackage{booktabs}
 \usepackage{graphicx}
\usepackage[T1]{fontenc}
\usepackage{xurl}

\usepackage[utf8]{inputenc}

\usepackage{microtype}
\usepackage{tipa}
\usepackage{comment}
\newcommand\todo[1]{{\color{red}TODO:#1}}

\usepackage[ruled,vlined,linesnumbered]{algorithm2e}

\newcounter{myexamp}
\newcommand{\numberedex}[1]{\noindent\refstepcounter{myexamp}\arabic{myexamp}\label{#1}}

\title{When your Cousin has the Right Connections: Unsupervised Bilingual Lexicon Induction for Related Data-Imbalanced Languages}

\name{Niyati Bafna,$^{1}$
  Cristina España-Bonet,$^{3}$
  Josef van Genabith,$^{2,3}$ \\
  {\large \textbf{Benoît Sagot},$^{1}$}
  {\large \textbf{Rachel Bawden}$^{1}$}}
\address{
  $^{1}$Inria, Paris, France\\
  $^{2}$Saarland University, Saarland Informatics Campus, Germany\\
  $^{3}$DFKI GmbH, Saarland Informatics Campus, Germany\\
  \texttt{
      nbafna1@jhu.edu,\{josef.van\_genabith,cristinae\}@dfki.de} \\
      \texttt{
      \{benoit.sagot,rachel.bawden\}@inria.fr
}}

\abstract{
Most existing approaches for unsupervised bilingual lexicon induction (BLI) depend on good quality static or contextual embeddings requiring large monolingual corpora for both languages. However, unsupervised BLI is most likely to be useful for low-resource languages (LRLs), where large datasets are not available. Often we are interested in building bilingual resources for LRLs against related high-resource languages (HRLs), resulting in severely imbalanced data settings for BLI. We first show that state-of-the-art BLI methods in the literature exhibit near-zero performance for severely data-imbalanced language pairs, indicating that these settings require more robust techniques. We then present a new method for unsupervised BLI between a related LRL and HRL that only requires inference on a masked language model of the HRL, and demonstrate its effectiveness on truly low-resource languages Bhojpuri and Magahi (with <5M monolingual tokens each), against Hindi. We further present experiments on (mid-resource) Marathi and Nepali to compare approach performances by resource range, and release our resulting lexicons for five low-resource Indic languages: Bhojpuri, Magahi, Awadhi, Braj, and Maithili, against Hindi. \\ \newline \Keywords{bilingual lexicon induction, low-resource, Indic languages} }

\begin{document}

\maketitleabstract
\section{Introduction}


Bilingual lexicons are a basic resource with varied uses, both in themselves, for dictionary building and language learning, as well as seeds for solving other problems in natural language processing (NLP), such as parsing \citep{zhao-etal-2009-cross,durrett-etal-2012-syntactic} and word-to-word and unsupervised machine translation \citep{irvine-callison-burch-2013-combining,thompson-etal-2019-hablexb}. 



While there is growing interest in unsupervised or minimally supervised bilingual lexicon induction (BLI), existing methods often depend on aligning monolingual word embedding spaces, assumed to be of good quality for both languages, and/or bilingual supervision \citep{artetxe-etal-2016-learning,artetxe-etal-2017-learninga,conneau-etal-2018-word,artetxe-etal-2018-robust,artetxe-etal-2018-generalizing,artetxe-etal-2019-bilingual}. However, extremely low-resource languages (LRLs) and dialects often lack good quality 
embeddings due to limited monolingual data, 
leading to very low or near-zero performance of alignment-based methods for these languages \citep{wada-etal-2019-unsupervised,eder-etal-2021-anchorbased}. 

This is the case for the under-researched Indic language continuum, which is the focus of this article (see Section~\ref{app:languages} for a description of the linguistic setup in India that motivates our work). We work with five extremely low-resourced Indic languages, Bhojpuri (\texttt{bho}), Magahi \texttt{mag}), Awadhi (\texttt{awa}), Maithili (\texttt{mai}), and Braj (\texttt{bra}), which are closely related to higher-resourced Hindi, and which have extremely limited resources, in terms of training data (<5M tokens of monolingual data) and embeddings, and even evaluation data. We demonstrate that state-of-the-art, alignment-based methods perform poorly in these settings, and introduce a new method for unsupervised BLI that performs much better. We aim to design methods that work well in characteristic data-scarce conditions, as well as generate resources for further work in these languages.  



Our main contribution is a novel unsupervised BLI method to address the typical scenario of the LRLs of the Indic continuum, i.e. for extremely LRLs that share significant overlap with a closely related HRL. We suppose that a masked language model (MLM) such as monolingual BERT \citep{devlin-etal-2019-berta} is available for the HRL and that we have some monolingual LRL sentences that contain unknown words. The method consists of building a lexicon iteratively by using the HRL MLM over LRL sentences to extract translation equivalents, and replacing learnt words in LRL sentences with HRL equivalents to make them more tractable for the HRL MLM for future unknown words (see Section~\ref{sec:method}). 

Given the lack of existing gold lexicons for our target languages (a frequent scenario for extremely LRLs), we create silver lexicons for Bhojpuri and Magahi created from parallel data, unfortunately unavailable for Awadhi, Maithili, and Braj. We also perform control experiments on Marathi and Nepali, two medium-resource languages more distantly related to Hindi with available gold lexicons, and discuss the performance of canonical methods and our proposed method on these languages, shedding light on what strategies are appropriate for differently-resourced language pairs. Our experiments indicate that current state-of-the-art methods are not suitable for low-resourced dialects, and methods that account for the data imbalance in the language pair, such as ours, may be more successful. We release our code, our generated lexicons for all languages (to our knowledge the first to be publicly released for all languages except Bhojpuri),\footnote{Although dedicated teams are working towards building resources for these languages \citeplanguageresource{mundotiya-etal-2021-linguistica-lr}, these resources (including evaluation resources) have not yet been made public as far as we know.} and our created silver evaluation lexicons for Bhojpuri and Magahi.\footnote{Code and resources available here: \url{https://github.com/niyatibafna/BLI-for-Indic-languages}.} See details of our released lexicons in Section~\ref{app:rel_lex}. 

Our motivation and method, while relevant to the 40+ resource-scarce languages of the Indic language family and other Indian languages, are also relevant to other linguistic systems with similar circumstances, i.e.~with a single high-or-medium resource language (usually a standard dialect), and several closely related dialects with lexical, morphological, and syntactic variation, written in the same script with or without orthographic standardization. This setup describes, for example, the Arabic continuum, the Turkic language continuum, and the German dialect system. 

\section{Linguistic Setup in India}
\label{app:languages}

\begin{table*}[t]
\small
\centering
\begin{tabular}{llllll}
\toprule
\textbf{Meaning} & \textbf{boy (nom)} & \textbf{sister (nom)} & \textbf{your (hon., fem. sing. obj)} & \textbf{told (completive)} & \textbf{(you) are going} \\
\midrule
\textbf{Hindi} & \textbf{\textipa{l\textschwa\:dk\textscripta:}} & \textbf{\textipa{b\textschwa h\textschwa n}} & \textbf{\textipa{a\textlengthmark pki\textlengthmark}} & \textbf{\textipa{b\textschwa \textsubbridge{t}\textlengthmark a\textlengthmark ja\textlengthmark/ k\textschwa \textlengthmark h} \textipa{lija\textlengthmark}} & \textbf{\textipa{d\textctz a\textlengthmark} \textipa{r\textschwa he\textlengthmark} \textipa{ho\textlengthmark}} \\ 
\textbf{Awadi} & \textipa{l\textschwa\:dk\textscripta:} & \textipa{b\textschwa hin} & \textipa{a\textlengthmark p\textschwa n} & \textipa{b\textschwa \textsubbridge{t}\textlengthmark a\textlengthmark v\textschwa \textsubbridge{t}} & \textipa{d\textctz a\textlengthmark \textsubbridge{t}} \textipa{\textschwa ha\textlengthmark i} \\
\textbf{Bhojpuri} & \textipa{l\textschwa ika\textlengthmark} & \textipa{b\textschwa hin} & \textipa{a\textlengthmark p\textschwa n} & \textipa{k\textschwa h\textschwa l} & \textipa{d\textctz a\textlengthmark \textsubbridge{t}} \textipa{ba\textlengthmark} \\
\textbf{Magahi} & \textipa{l\textschwa i\textlengthmark ka\textlengthmark} & \textipa{b\textschwa hin} & \textipa{\textschwa p\textschwa n} & \textipa{k\textschwa h\textschwa lie\textlengthmark} & \textipa{d\textctz a\textlengthmark} \textipa{h\textschwa i} \\
\textbf{Maithili} & \textipa{l\textschwa\:dk\textscripta:} & \textipa{b\textschwa hin} & \textipa{\textschwa ha\textlengthmark nk} & \textipa{k\textschwa h\textschwa lhu\textsuperscript{n}} & \textipa{d\textctz a\textlengthmark} \textipa{r\textschwa h\textschwa l} \textipa{\textschwa\textteshlig\textsuperscript{h} i} \\

\bottomrule
\end{tabular}
\caption{Examples of cognates. Since the Devanagari script is phonetically transparent, phonetic similarity is visible both in IPA and in Devanagari (not shown).}
\label{tab:cognates}
\end{table*}

\begin{table*}[!ht]
    \centering
    \includegraphics[scale=1]{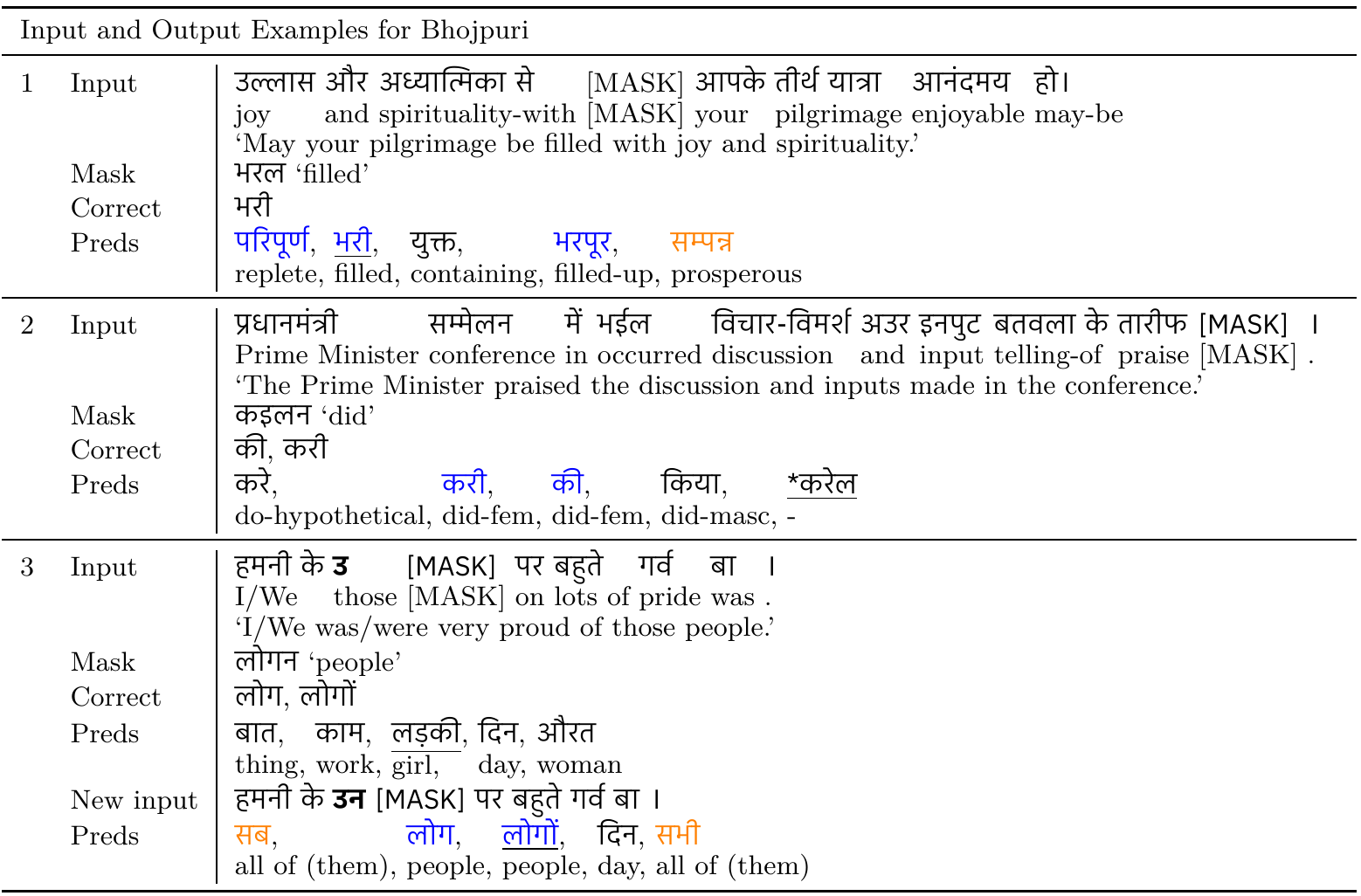}
    \caption{I/O examples, shown for Bhojpuri and the outputs of \textsc{Basic}. \textbf{Input}: Target language text with an unknown masked word. \textbf{Mask}: the original masked target language word. \textbf{Correct}: Acceptable Hindi equivalents given in the silver lexicon. \textbf{Preds}: The top predictions made by the mask-filling model. We choose $5$ representative examples here(in practice we use the top $30$). The underlined prediction is the one that has the lowest normalized edit distance with the original target language word, and would therefore be chosen in the \textsc{Basic} approach as the top candidate. \textcolor{blue}{Blue} highlighted predictions are correct equivalents, even if not included in silver lexicon. \textcolor{orange}{Orange} highlighted examples fit the mask but are not equivalents of the masked target word. * indicates a non-word in Hindi. \textbf{New input}: Input with known source word replaced with target equivalent (indicated in bold).}
\label{tab:method_examples}
\end{table*}


India has around 15-22 languages that are medium-to-high-resource, such as Hindi, Marathi, and Tamil, but dozens of other languages and dialects that are extremely low-resourced, with very little monolingual data (<5M tokens), and no other resources, such as Marwadi, Tulu, Dogri, and Santhali. These languages are often closely related to at least one high-resource language (HRL), meaning that they share morphosyntactic properties as well as a high number of cognates \citep{jha-2019-exploring,mundotiya-etal-2021-linguistica} (see Table~\ref{tab:cognates} for examples). They often have no official status in the regions where they are spoken, and therefore do not have concerted funding efforts for data collection or research. Even when such efforts do exist,\footnote{See \url{https://data.ldcil.org/text}.} the collected corpora are rarely of the magnitude at which static or contextual embeddings can be well-estimated. While the actual number of distinct dialects and languages spoken in India is contested, people self-reported about 576 such ``mother tongue'' dialects in the latest census,\footnote{See \url{https://www.outlookindia.com/national/explained-what-is-mother-tongue-survey-and-its-importance-in-preserving-india-s-linguistic-data-news-235854}.} which were then grouped into around 121 languages. Only 22 of these languages have official status (i.e.~they are either the official language of some state/union territory, or have national cultural significance), and are therefore accorded funds for the development of resources. 

Therefore, although some studies in the literature question the real use case for entirely unsupervised BLI \citep{vulic-etal-2019-wea}, since it is ``easy'' to collect a small bilingual lexicon, we argue that situations such as these, where there is a large number of languages to build support for, and where efforts in data collection and annotation for individual languages are restricted by the availability of funds, do constitute genuine application scenarios for unsupervised BLI. 

Furthermore, we focus on a scenario where the two languages in question are closely related. This is because for most of the low-resource languages in the Indian context cited above, we can usually find a linguistic neighbour that is relatively well off, usually one of India's $22$ scheduled languages.\footnote{Of course, there are exceptions to this observation; naturally, this is not true for language isolates such as Burushaski, spoken in Pakistan, or for the small minority of Austro-Asiatic langauges such as Mundari, spoken in India, which do not have a single (Indian) high-resource sister language.} In general, when building resources for a given low-resource dialect or language, it is likely that the standard variant of that dialect, or the HRL closest to it, will have large enough corpora available to build a good quality MLM. We target our efforts to these situations.

\section{Related Work}


Recent years have seen interest in unsupervised BLI \citep{haghighi-etal-2008-learning,artetxe-etal-2016-learning,artetxe-etal-2017-learninga,conneau-etal-2018-word,artetxe-etal-2018-robust,artetxe-etal-2018-generalizing,artetxe-etal-2019-bilingual}, allowing the possibility of BLI for LRLs. Most unsupervised approaches, notably \textsc{MUSE} \citep{conneau-etal-2018-word} and \textsc{VecMap} \citep{artetxe-etal-2016-learning,artetxe-etal-2017-learninga,artetxe-etal-2018-robust,artetxe-etal-2018-generalizing} are based on training static embeddings from large monolingual corpora \citep{mikolov-etal-2013-efficient,bojanowski-etal-2017-enriching}, and aligning the embeddings using linear or non-linear mappings, using an initial seed \citep{xing-etal-2015-normalized,artetxe-etal-2016-learning}. 

Recent works have also looked at using contextual embeddings or BERT-based models \citep{peters-etal-2018-deepa,devlin-etal-2019-berta,ruder-etal-2019-survey} for BLI. \citet{gonen-etal-2020-it} induce word-level translations by directly prompting mBERT \citep{devlin-etal-2019-berta}. \citet{yuan-etal-2020-interactive} present a human-in-the-loop system for BLI in four low-resource languages, updating contextual embeddings with the help of annotations provided by a native speaker.  \citet{zhang-etal-2021-combining} present \textsc{CSCBLI}, a method that uses a ``spring network'' to align non-isomorphic contextual  embeddings, and interpolates them with static embeddings to estimate word similarities, showing superior results to other methods using contextual embeddings, notably BLISS \citep{patra-etal-2019-bilingual}. These approaches rely on parallel data or large monolingual corpora for good quality contextual embeddings. However, for low-resource languages, contextual embeddings from both monolingual and multilingual models are known to be unreliable \citep{wu-dredze-2020-are}.

Later works show the failings of the above approaches in low-resource settings \citep{adams-etal-2017-crosslingual,kuriyozov-etal-2020-crosslingual,chimalamarri-etal-2020-morphological,eder-etal-2021-anchorbased} and propose alternative training strategies such as joint training of static embeddings \citep{woller-etal-2021-not,bafna-etal-2022-combininga}, and multilingual embeddings from LSTM-based models \citep{wada-etal-2019-unsupervised}. However, these works either address a higher resource range (>15M tokens), use bilingual lexicons as seeds, or show low scores ($\approx$30 precision@5) for unsupervised BLI. In general, there is a paucity of attention given to setups where there is a severe resource imbalance between the two languages of the BLI pair, despite this being a very typical real-world scenario.


\section{Method}
\label{sec:method}
Our method is intended for a closely related HRL (source) and LRL (target) pair, written in the same script, and given that we can train or already have a good quality monolingual MLM for the HRL. The main idea is that if we mask an unknown word in the LRL sentence, feed the masked LRL sentence to the HRL MLM, and ask the HRL MLM to propose candidates for the masked LRL word, the HRL MLM should have access to enough contextual cues due to shared vocabulary and syntax to propose meaningful HRL candidates for the masked word. This potentially gives us translation equivalence between the original LRL word and the best scoring proposed HRL candidate. We proceed in an iterative manner, growing the lexicon from equivalents gained from each processed sentence, and using learned equivalents in the lexicon to replace known LRL words with HRL equivalents to process future sentences. 

Starting with an empty HRL-LRL bilingual lexicon, we perform the following steps to update our lexicon iteratively, explained in further detail below, and shown in Algorithm~\ref{alg:basic}: (i)~we choose an input, consisting of an LRL sentence, and a source LRL word occurring in it, (ii)~we replace known words in the input sentence by HRL equivalents using the current state of the lexicon, in order to make the sentence more HRL-like, (iii)~the resulting sentence is passed to the HRL MLM to obtain HRL candidate suggestions for the masked LRL word, (iv)~we use a reranking heuristic to choose the best HRL candidate, if any, and (v)~we update the lexicon if we have found a new equivalent pair.

\subsection{Choosing \textit{(sentence, word)} pairs to process} Intuitively, the chance of the HRL MLM giving accurate translation equivalents for the (LRL) word is higher if the LRL sentence is more easily ``comprehensible'' to the HRL MLM, or if the LRL sentence already has several HRL words in it. Therefore, we aim to first process words in sentences that have a higher concentration of known words, where known words are either shared vocabulary or words that are already in the current state of our lexicon. These words are replaced by their HRL equivalents before the sentence is passed to the HRL MLM.\footnote{We mask whole words and accept single token responses (as the default) from the MLM. In practice, this does not pose a big problem, since the HRL MLM tokenizer has a large vocabulary size (52000): 86\% and 81\% in the Hindi side of the Bhojpuri and Magahi silver lexicons respectively are preserved as single tokens. We leave it to future work to handle multi-word terms.} We maintain a priority list of \textit{(sentence, word)} pairs based on the percentage of known words in the sentence and update the list after every batch of sentences 
based on new learned translations.\footnote{Specifically, the priority list is created from the \textit{(sentence, unk\_word)} pairs by first sorting them by the number of times each instance has previously been processed, and then by the percentage of other unknown words in the sentence, both in ascending order.}

\subsection{Reranking} The HRL MLM may propose valid candidates for the masked token that are not translation equivalents to the LRL source word; typically, there may be a wide range of reasonable possibilities for any masked word. Therefore, we rerank the returned HRL candidates based on orthographic closeness to the masked LRL word. Our use of orthographic closeness as the basis of our rerankers is motivated by the high percentage of orthographically similar cognates, borrowings, and spelling variants in the vocabulary of these languages with respect to each other (shown by \citet{jha-2019-exploring} for Maithili and Hindi). Note that minimum normalized edit distance as a stand-alone approach, i.e.~positing the orthographically closest HRL word as a translation equivalent for any LRL word, performs badly for various reasons \citep{bafna-etal-2022-combininga}. We compare two rerankers, \textsc{Basic} and \textsc{Rulebook}.


\subsubsection{Basic}
In the \textsc{Basic} approach, we simply use normalized orthographic similarity (computed using Levenshtein distance) between the candidate and the original masked word. This reranker considers all character substitutions equally costly.

\subsubsection{Rulebook}

We may see from discovered cognate pairs that certain character transformations are very common (corresponding to regular sound change, or systematic differences in orthographic conventions), and so should be less costly than others. Similarly, different language pairs may have different preferences for cheap or costly character substitutions. 

In the \textsc{Rulebook} variant, we use \citeauthor{bafna-etal-2022-combininga}'s (\citeyear{bafna-etal-2022-combininga}) iterative expectation-maximization (EM) method to learn a custom edit-distance matrix for the source and target character sets. This custom edit-distance matrix is used as an orthographic reranker for our approach (lines~6-9 in Algorithm~\ref{alg:basic}). 

The idea of this reranker is to iteratively optimize character substitution probabilities from the source to target character set in ``known'', or hypothesized, cognate pairs, while simultaneously learning new cognate pairs by reranking candidates suggested by the HRL MLM, using the current state of the substitution probabilities.

\paragraph{Setup}
Let $\chi_s$ and $\chi_t$ represent the sets of characters on the source (LRL) and target (HRL) sides, respectively. We define a scoring function, $S(c_i, c_j)$ that provides a score for replacing a character $c_i \in \chi_s$ with $c_j \in \chi_t$. Insertions and deletions are considered special cases of replacement, where a null character is introduced or replaced. For a given source set character, $S$ is modelled as a transformation probability distribution over $\chi_t$. Initially, the probabilities in $S$ are assigned to favor self-transformations (typically set to 0.5), and the remaining probability mass is evenly distributed among other characters.

At any given iteration, we can calculate the score for a source-target character substitution, viewed as a conditional probability:
\begin{equation}
 S(c_i, c_j) = \frac{C(c_i, c_j)}{T(c_i)}    
\end{equation}
Here, $C(a, b)$ is the number of times we have seen $a$ $\rightarrow$ $b$, and $T(a)$ is the total number of times we have seen $a$ on the source side.


\paragraph{EM Steps} for \textsc{Rulebook}. 

\textit{1) Expectation step.} Given a list of top $k$ candidates for a given source word $s$: for each candidate pair $(s, t)$, we find $Ops(s, t)$, which is the \textit{minimal list} of the operations we need to perform to get from $s$ to $t$. Each member in $Ops$ is of the type $(c_i, c_j)$. Note that we also want to estimate $S(a, a)$ $\forall$ $a$, and so we also use a ``retain'' operation, for characters that remain the same. 
The score for the pair $(s, t)$ is computed as:
\begin{equation}
\zeta(s, t) = -\sum_{(a,b) \in Ops} log(S(a, b)),    
\end{equation}

\noindent
where the lower the $\zeta$ the more probable it is that a pair is equivalent.
For a given $s$, we can then always find the word that is the most probable equivalent as 
$t_{best} = argmin_{t_i \neq s}(\zeta(s, t_i))$ (line~6 in Algorithm~\ref{alg:basic}). We then add $(s, t_{best})$ to our learned lexicon (line~8).

\textit{2) Maximization step.} We update the model parameters based on the newly identified equivalents in the previous step (line~9 in Algorithm~\ref{alg:basic}). This is done by increasing the counts of all observed edit distance operations:
$$ C(a, b) := C(a, b) + 1 ~~~~\forall (a, b) \in Ops(s, t) $$
$$ ~~~~~T(a) := T(a) + 1 ~~~~~~~\forall (a, b) \in Ops(s, t) $$
We disallow updates for $s=t$ (i.e.~identical words) in the training phase, to mitigate exploding self-transform probabilities.

\subsection{Multiple passes over the input} Once all $(sentence, word)$ pairs have been processed once (or $n$ times), we reprocess them (for an $(n+1)^{th}$ pass) in the hope of gaining more accurate translations, as previously unknown neighbour words may have been learned in the meantime. 

\subsection{Hyperparameters} We use a minimum normalized orthographic similarity threshold of 0.5 (see line~7 of Algorithm~\ref{alg:basic}). This threshold was heuristically chosen. We set the maximum number of passes to 3, meaning that the algorithm terminates if all unknown words have been processed 3 times. We found in our initial experiments that the algorithm yields very few or no new words in further passes. This also serves as a terminating condition (line~1 in Algorithm~\ref{alg:basic}). 

\subsection{Examples}
\label{app:method_ex}
We give examples of inputs and outputs of our method in Table~\ref{tab:method_examples}, illustrating the outputs for \textsc{Basic}. As we see, the Hindi BERT is fairly good at giving reasonable Hindi candidates for the masked Bhojpuri, although, naturally, these candidates may not be equivalents of the masked word, as shown for the top candidates in rows~2 and~3. Applying reranking based on orthographic similarity solves this problem to a large extent, serving to identify translation equivalents from among given candidates.

We also see an example (row~3) where replacing a Bhojpuri word with its Hindi equivalent in the input sentence helps the Hindi MLM to produce more reasonable Hindi candidates for the masked word.

\begin{algorithm}[htbp]

\small
\SetKwComment{Comment}{// }{}
\caption{\textsc{Basic} and \textsc{Rulebook}}
\label{alg:basic}

    \While{not $terminating\_condition$}{
    sent, word $\gets$ \textit{chooseLRLExample}()\;
      sent $\gets$ \textit{replaceKnownWords}(sent, lexicon); \\
      sent $\gets$ \textit{maskWord}(sent, word); \\
      preds $\gets$ \textit{HRLBertMaskFill}(sent); \\
      best $\gets$ \textit{argmax}(\textit{orthSim}(word, preds)); \\
      \If{\textit{orthSim}(best, word) > threshold}{
       lexicon $\gets$ \textit{updateLexicon}(word, best); \\
       \tcc{Next step only for \textsc{rulebook}}
       \textit{updateOrthSimParams}(word, best)
      }
      }
\end{algorithm}

\begin{table*}
    \centering
    \includegraphics[scale=0.3]{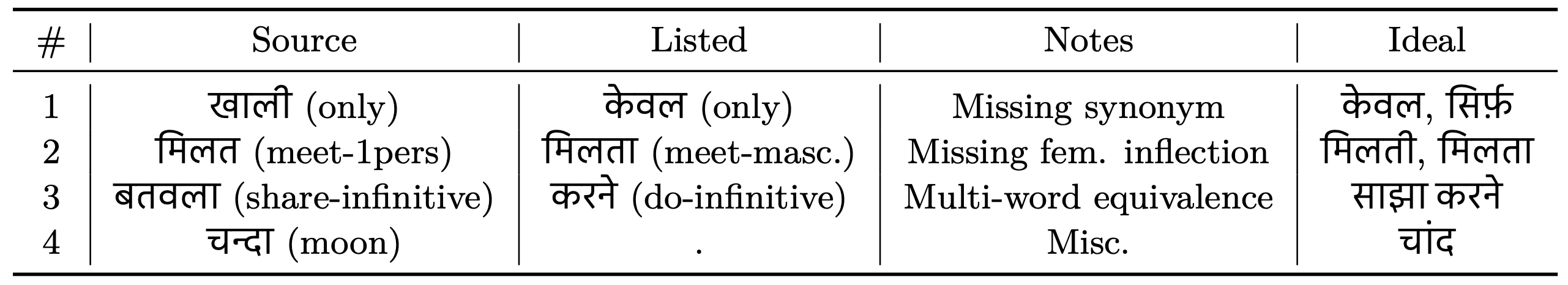}

    \caption{Types and examples of faults in the silver lexicon.}
    \label{tab:silver_lex_ex}
\end{table*}
\begin{table}
\centering
\small
\begin{tabular}{lrrr}
\toprule
Target & \#Tokens & Lexicon & Silver lexicon \\
lang. & & size & size \\
\midrule
awa &  0.17M  & 10462  & - \\
bho &  3.09M   & 21983 & 2469 \\
bra &  0.33M & 10760  & - \\
mag &  3.16M & 30784 & 3359 \\
mai &  0.16M  & 12069  & - \\
mar* &  551.00M & 36929 & - \\
nep* & 110.00M & 22037 & - \\
\bottomrule
\end{tabular}
\caption{Monolingual data sizes in tokens, and sizes of our released lexicons (created using our method), and released silver lexicons (from parallel data) for Bhojpuri and Magahi. *High-quality gold bilingual lexicons already exist for these languages.}
\label{tab:lex_sizes}

\end{table}

\section{Experimental Settings}
\label{sec:data}

\paragraph{Monolingual Data}

We use monolingual data from the LoResMT shared task \citeplanguageresource{Ojha_Malykh_Karakanta_Liu_2020} for Bhojpuri and Magahi, and the VarDial 2018 shared task data \citeplanguageresource{zampieri-etal-2018-proceedings} for Bhojpuri, Awadhi and Braj. For Bhojpuri, we additionally use the BHLTR project \citeplanguageresource{ojha-2019-englishbhojpuria}. We use the BMM corpus 
\citeplanguageresource{mundotiya-etal-2021-linguistica-lr} and the Wordschatz Leipzig corpus \citeplanguageresource{goldhahn-etal-2012-buildinga} for Maithili. For Marathi and Nepali, we use large-scale monolingual corpora made available by IndicCorp \citeplanguageresource{kakwani-etal-2020-indicnlpsuitea} and \citelanguageresource{jxrd-d245-20} respectively. See Table~\ref{tab:lex_sizes} for monolingual data sizes.

\paragraph{Model} We use the MuRIL model and tokenizer \citep{khanuja-etal-2021-muril} as our HRL MLM for Bhojpuri, Magahi, Awadhi, Maithili and Braj; we use the Hindi BERT and associated tokenizer given by \citet{joshi-2023-l3cubehindbert} for Marathi and Nepali.\footnote{We need a HRL model that has not seen target data; while MuRIL is a good choice for low-resource dialects because it is multilingual and may benefit from knowledge of other related Indic languages, it cannot be used for Marathi and Nepali because these languages are included in its pretraining data.}

\paragraph{Baselines}
We compare our approaches against semi-supervised \textsc{VecMap} approach with CSLS \citep{artetxe-etal-2018-generalizing,artetxe-etal-2018-robust}, using identical words as seeds, with 300-dimensional fastText embeddings \citep{bojanowski-etal-2017-enriching}.\footnote{Using $100$ dimensions gives similar results.} We also choose \textsc{CSCBLI} \citep{zhang-etal-2021-combining} as a representative of methods using contextual representations, hypothesizing that the ensemble of static and contextual embeddings may perform better than \textsc{VecMap}. Finally, we report results for a trivial baseline \textsc{ID}, the identity function, representing vocabulary overlap.

\paragraph{Evaluation Data}
Given the lack of gold lexicons between Hindi and our LRLs, we create silver lexicons instead from parallel data. We use FastAlign with GDFA \cite{dyer-etal-2013-simple} to extract word alignments from existing gold Bhojpuri--Hindi and Magahi--Hindi parallel data ($\approx$500 sentences per language) \citeplanguageresource{ojha-2019-englishbhojpuria}.\footnote{We were unable to find a reasonable quantity of publicly available parallel data for Awadhi, Braj, or Maithili, and so could not perform evaluation for these languages.} We use the two best candidates per source word in the resulting silver lexicons as valid translations.\footnote{This is an empirical choice from eyeballing the resulting silver lexicons; at least one and usually both of the first two translations are valid.} This yields 2,469 and 3,359 entries for Bhojpuri and Magahi respectively. We report the manually evaluated quality of the silver lexicons in the following paragraph. For Marathi and Nepali, we use existing gold parallel lexicons against Hindi, taken from IndoWordNet \citeplanguageresource{kakwani-etal-2020-indicnlpsuitea}, manually aligned to the Hindi WordNet. We obtain lexicons with 35,000 and 22,000 entries for Marathi and Nepali respectively.

\paragraph{Manual Evaluation of Silver Lexicons}
\label{app:man_eval_silver_lex}

We perform a manual evaluation of our silver lexicons, in order to judge the credibility of the reported results for our methods for Bhojpuri and Magahi. We manually examine 150 entries in the automatically created Bhojpuri lexicon, and find that 90\% of entries are satisfactory, i.e.~they list accurate Hindi equivalents of Bhojpuri words. We observe a few general problems with the lexicon, and list representative examples in Table~\ref{tab:silver_lex_ex}: 

\begin{itemize}
    \item Missing common synonyms, e.g. in row 1 of Table~\ref{tab:silver_lex_ex}. This kind of error results in underestimation of precision scores for all approaches. 
    \item Problems with correctly equating inflections, missing feminine inflections, e.g.~row 2. A natural problem arising from differences in morphological systems of the source and target language is that inflected verbs can be difficult to match cross-lingually. This results in missing equivalents of a given inflected form. For example, while genderless verbs in Bhojpuri should ideally be listed with the corresponding masculine and feminine verbs in Hindi, we observe that they are often missing one gender inflection, usually the feminine one. Similarly, not all possible target inflectional variants of a source inflection are listed for each verb entry. 
    \item Multi-word equivalences lead to errors. For example, in row~3, the single-word Bhojpuri source verb has a noun-light verb complex equivalent in Hindi (consisting of two words, literally meaning ``sharing do''), and the silver lexicon lists the light verb (``do'') as the target translation. This is also observed in the case of other verb equivalences, where one of the languages using multiple tokens to express an inflection, leading to incorrect matches in the silver lexicon. 
    \item Miscellaneous errors. The lexicon 
    contains some entirely incorrect equivalents (8.76\%), due to word alignment errors, e.g.~row~4. 
    
\end{itemize}

Note that we only mark entries as wrong if the listed equivalents are inaccurate, and so faults such as missing synonyms and inflections, which affected 7.33\% of the sample we examined, are not represented in the error percentage reported.





\section{Results and Discussion}

\begin{table*}[t]
\centering\small
\begin{tabular}{llrrrrrrrr}
\toprule
& & \multicolumn{2}{c}{bho} & \multicolumn{2}{c}{mag} & \multicolumn{2}{c}{mar} & \multicolumn{2}{c}{nep} \\
\midrule
& Method & P@2 & NIA & P@2 & NIA & P@2 & NIA & P@2 & NIA \\
\midrule
Baselines & ID & 37.3 & 0.0 & 39.9 & 0.0 & 27.5 & 0.0 & 21.2 & 0 \\
& VecMap+CSLS & 0.0 & 0.0 & 1.2 & 0.6 & 42.4 & \textbf{26.7} & 0.0 & 0.0 \\
& CSCBLI & 0.0 & 0.0 & 2.0 & 0.5 & 0.0 & 0.0 & 0.0 & 0.0 \\
\midrule
Ours & Basic & 61.0 & \textbf{18.1} & 65.2 & \textbf{18.8} & \textbf{80.9} & 2.8 & \textbf{87.6} & \textbf{8.2} \\
& Rulebook & \textbf{61.5} & 15.1 & \textbf{65.4} & 17.4 & 80.6 & 1.72 & \textbf{87.6} & 6.0 \\
\bottomrule
\end{tabular}
\caption{Performance of the methods, given by Precision@2 (P@2) and accuracy of non-identical predictions (NIA).}
\label{tab:results}
\end{table*}

\begin{table*}
    \centering
\resizebox{\linewidth}{!}{   \includegraphics{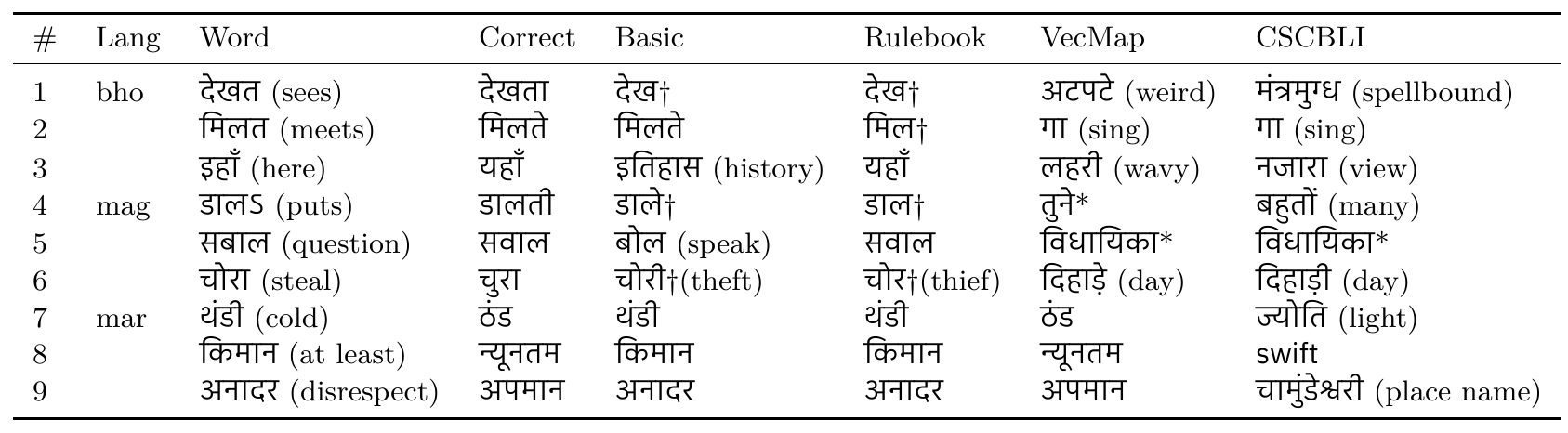}}
    \caption{Predictions made by different approaches. Meanings are provided for the first occurrence of the word. * indicates a non-word and \textdagger{} a prediction in the wrong inflectional/derivational form of the target.}
    \label{tab:predictions}
\end{table*}


We report precision@2 and accuracy on non-identical predictions (NIA) in Table~\ref{tab:results}.\footnote{We also report P@\{1,3,5\} in Appexdix~\ref{app:p135}; these results show similar trends.} \textsc{NIA} is calculated by taking all non-identical predictions in the top 2 predictions per word,  and reporting the percentage of those predictions that were marked correct by the evaluation lexicons. We report this metric because precision@2 may be inflated by ``easy'' identical word predictions.


\paragraph{Baselines} Table~\ref{tab:predictions} provides examples of the performance of these approaches. \textsc{VecMap} performs well for Marathi: we provide examples where it predicts correct equivalents for rare words (row~8), non-cognates (row~9), as well as frequent words (row~7). However, for Nepali, Bhojpuri, and Magahi, both \textsc{VecMap} and \textsc{CSCBLI} make seemingly random wrong predictions on almost all words (rows~1,~2, and~6), with near-zero performance, probably due to the low quality of static and contextual embeddings for the LRLs.
\textsc{CSCBLI} also fails for Marathi, indicating that the Marathi contextual embeddings may still be of poor quality or that the approach may not generalize well to untested language pairs. While the failure of these baselines for Nepali is surprising, it can perhaps be explained by the fact that Nepali has about five times less data than Marathi, and less lexical overlap with Hindi. 


\paragraph{Our methods}  Our \textsc{Basic} and \textsc{Rulebook} approaches outperform \textsc{ID} by more than 20 accuracy points for all languages. \textsc{Rulebook} gains very little, if at all, over \textsc{Basic}, but  \textsc{Rulebook} has an edge when it comes to predicting cognates with common sound correspondences (see row~5). We observe that these approaches are reasonably successful for Bhojpuri and Magahi on cognate verbs and common nouns, but fail on syntactic words and postpositions (row~3 for \textsc{Basic}), and may be confused by unrelated words with chance orthographic similarity even for common words (row~5 for \textsc{Basic}). Furthermore, these approaches often predict incorrect inflections of the correct verbal/noun stem (we count these predictions as wrong), as in rows~1~and~4. Although \textsc{Basic} and \textsc{Rulebook} perform with high accuracy for Marathi and Nepali, their \textsc{NIA} is extremely low, indicating that they serve mainly to identify or ``sieve'' out vocabulary overlap. We see that the candidates proposed by the Hindi MLM are often in fact Marathi/Nepali words, indicating that it has seen some Marathi/Nepali data (due to corpus contamination and/or code-mixing) and is capable of performing mask-filling for Marathi/Nepali.

\paragraph{Manual Evaluation of Generated Lexicons}
We manually examine errors in the non-identical predictions of \textsc{Basic}, looking at 60 randomly chosen non-identical 
Bhojpuri predictions.\footnote{One of the authors is a native speaker of Hindi but not Bhojpuri. We segregate translation equivalents into error categories using (self-made) inflection tables inferred from the silver lexicons, as well as cognate knowledge from Hindi native speaker knowledge.} 
We find that 31.7\% of predictions are correctly inflected equivalents, as opposed to 18.1\% given by the \textsc{NIA} quantitative evaluation. The underestimation is caused by missing synonyms in the silver lexicon. Furthermore, 25\% are incorrectly inflected cognates of the source word, and the rest are unrelated words.



\paragraph{How useful is reranking by orthographic distance?} We also ran the \textsc{Basic} approach without reranking with orthographic distance, i.e.~we simply pick the top candidate suggested by the HRL mask-filling model as an equivalent. This approach is clearly worse than the standard \textsc{Basic} approach (with reranking), performing at only 3.03\% \textsc{NIA} for Bhojpuri and 4.04\% \textsc{NIA} for Magahi (approximately -15 and -14 percentage points compared to \textsc{Basic} for Bhojpuri and Magahi respectively, as shown in Table~\ref{tab:results}). However, this approach can still identify and capture identical vocabulary. 

\paragraph{Variants} 

We experimented with minor variants of the \textsc{Rulebook} update mechanisms to see if they result in boosts to performance. We tried disallowing updates for the null character, since we found that a large probability mass iteratively accumulates in the null character (or for deletion). We also incorporated a change in the original algorithm, whereby we made updates to the custom edit distance matrix based on the optimal list of substitutions as per the current state of the edit distance matrix, rather than choosing a minimal length path at random (with each substitution counted as length $1$) from the source to the target word when several exist. However, these variants result in very minor improvements or even slight degradations to performance, and we do not report these results.

\section{Details of released lexicons}
\label{app:rel_lex}

We make our bilingual lexicons publicly available under a CC BY-NC 4.0 license for Bhojpuri, Magahi, Awadhi, Braj, and Maithili, and also release our created silver evaluation lexicons for Bhojpuri and Magahi under the same license. These are the first publicly available bilingual lexicons all these languages except Bhojpuri, to the best of our knowledge. The sizes of the released lexicons for each target language are provided in Table~\ref{tab:lex_sizes}. Note that while we also release our generated lexicons for Marathi and Nepali, large high quality gold bilingual lexicons already exist for these languages (see Section~\ref{sec:data}) and should be used instead of ours; we are mainly interested in creating resources for the low-resource languages.



\section{Conclusion}

We introduce a novel method for unsupervised BLI between a related LRL and HRL, which only requires a good quality MLM for the HRL. This addresses an important gap in the existing literature, which often relies on good quality embeddings for both languages. Our method shows superior performance on two low-resource languages from the Indic continuum, against near-zero performances of existing state-of-the-art methods. We perform control experiments for two more distantly related Indic languages, and release resulting bilingual lexicons for five truly low-resource Indic languages. 

\section*{Limitations}

The applicability of our method is restricted to low-resource languages that are related to a high-resource language. As the \textsc{Basic} and \textsc{Rulebook} method are directly dependent on orthographic distance between translation pairs, they are only useful for identifying cognate equivalents, borrowings, or alternate spellings in the source and target language. We also clarify that our method is not intended for mid-to-high resourced language pairs (such as Marathi--Hindi), where canonical state-of-the-art methods such as \textsc{VecMap}  work more robustly, specifically on non-identical word equivalents. Our method therefore has a specific (although important) target scenario, i.e.~it is a simple method to build bilingual lexicons for severely under-resourced languages leveraging the resources of a closely related high-resource language, given that state-of-the-art methods fail in these settings. Note that we also only deal in the entirely unsupervised scenario in keeping with typical conditions for our target languages (see Section~\ref{app:languages}), and leave it to future work to improve these methods with a little supervision from bilingual lexicons, possibly obtained from parallel data.

Another limitation of our work is that we were not able to provide true native speaker evaluation for the resulting target language lexicons, instead providing evaluation by the first author (Hindi native speaker) relying on knowledge of shared cognates, the morphology of the target language, and inflection tables. We provide examples in Table~\ref{tab:predictions} and Table~\ref{tab:method_examples}, and release the automatically created as well as silver lexicons. Finally, our method is only capable of providing single token (HRL) matches to the masked (LRL) whole word. As discussed in Section~\ref{sec:method}, this problem does not affect the large majority of cases. We leave it to future work to extend our idea to handle multi-token words and multi-word expressions using, for example, span-filling language models \citep{donahueetal2020enabling}.


\section*{Ethics Statement}

Our work is driven by the aim to boost NLP for severely under-resourced languages of the Indic language belt, as well as contribute a method that may be relevant to other language families with a similar linguistic and  resource setup. 
Our method relies on the predictions of language models for the high-resource language and is therefore fallible to general ethical issues with such models, including caste, religion, and gender biases shown to be exhibited by such models \citep{malik-etal-2022-socially}. 

\section*{Acknowledgements}

This work was partly funded by the last two authors' chairs in the PRAIRIE institute funded by the French national agency ANR as part of the ``Investissements d'avenir'' programme under the reference ANR-19-P3IA-0001. First and second authors are supported  by the Deutsche Forschungsgemeinschaft (DFG, German Research Foundation) ---Project--ID 232722074--- SFB 1102. Second and third authors are supported by the EU project LT-Bridge (GA952194).


\nocite{*}
\section{Bibliographical References}\label{sec:reference}

\bibliographystyle{lrec-coling2024-natbib}
\bibliography{main}

\section{Language Resource References}
\label{lr:ref}
\bibliographystylelanguageresource{lrec-coling2024-natbib}
\bibliographylanguageresource{languageresource}
\newpage
\appendix
\begin{table}[ht]
\centering
\small
\begin{tabular}{lrrrrrr}
\toprule
& \multicolumn{3}{c}{bho} & \multicolumn{3}{c}{mag} \\
\midrule
& P@1 & P@3 & P@5 & P@1 & P@3 & P@5 \\
\midrule
VecMap+CSLS & 0 & 0 & 0 & 1.2 & 1.2 & 1.2 \\ 
\midrule
Basic & 58.1 & 61 & 61 & 62.5 & 65.1 & 65.1 \\ 
Rulebook & \textbf{59.1} & \textbf{61.6} & \textbf{61.6} & \textbf{63} & \textbf{65.4} & \textbf{65.4} \\ 
\bottomrule
\end{tabular}
\caption{P@\{1,3,5\} for \texttt{bho} and \texttt{mag}.}
\label{tab:app_prec}
\end{table}

\begin{table}[ht]
\centering\small
\begin{tabular}{lrrrrrr}
\toprule
& \multicolumn{3}{c}{bho} & \multicolumn{3}{c}{mag} \\
\midrule
& NIA@1 & NIA@3 & NIA@5 & NIA@1 & NIA@3 & NIA@5 \\
\midrule
VecMap+CSLS & 0&	0&	0&	0.6&	0.6&	0.6 \\ 
\midrule
Basic & \textbf{23.7}&	\textbf{17.1}&	\textbf{17.1}&	\textbf{27.2}&	\textbf{18.4}& \textbf{18.2}  \\ 
Rulebook & 20.2&	14.4&	14.5&	23.1&	17&	16.9 \\ 
\bottomrule
\end{tabular}
\caption{NIA@\{1,3,5\} for \texttt{bho} and \texttt{mag}.}
\label{tab:app_nia}
\end{table}

\section{Additional Results}
\label{app:p135}
We report P@{1,3,5} in Table~\ref{tab:app_prec} and $\textsc{NIA}$@{1,3,5} in Table~\ref{tab:app_nia}. We see that both \textsc{Basic} and \textsc{Rulebook} approaches do not benefit from considering more than 3 best answers. In general, we see the same relative trend as in Table~\ref{tab:results}.

\end{document}